\documentclass[10pt,twocolumn,letterpaper]{article}

\usepackage{iccv}
\usepackage{times}
\usepackage{epsfig}
\usepackage{graphicx}
\usepackage{amsmath}
\usepackage{amssymb}

\usepackage[pagebackref=true,breaklinks=true,letterpaper=true,colorlinks,bookmarks=false]{hyperref}

\iccvfinalcopy 

\ificcvfinal\fi

\usepackage{epsfig}
\usepackage[labelformat=simple]{subfig}

\begin{document}





\title{From Image to Text Classification: A Novel Approach based on Clustering Word Embeddings}


\author{Andrei M. Butnaru and Radu Tudor Ionescu
\\
University of Bucharest, 14 Academiei, Bucharest, Romania}

\maketitle

\begin{abstract}
In this paper, we propose a novel approach for text classification based on clustering word embeddings, inspired by the \emph{bag of visual words} model, which is widely used in computer vision. After each word in a collection of documents is represented as word vector using a pre-trained word embeddings model, a k-means algorithm is applied on the word vectors in order to obtain a fixed-size set of clusters. The centroid of each cluster is interpreted as a \emph{super word embedding} that embodies all the semantically related word vectors in a certain region of the embedding space. Every embedded word in the collection of documents is then assigned to the nearest cluster centroid. In the end, each document is represented as a \emph{bag of super word embeddings} by computing the frequency of each super word embedding in the respective document. We also diverge from the idea of building a single vocabulary for the entire collection of documents, and propose to build class-specific vocabularies for better performance. Using this kind of representation, we report results on two text mining tasks, namely text categorization by topic and polarity classification. On both tasks, our model yields better performance than the standard \emph{bag of words}.
\end{abstract}


\vspace*{-5pt}


\section{Introduction}
\label{main}

With the recent exponential growth of the Internet, there is more and more data that requires efficient processing methods for storing and extracting relevant information. This data is usually unstructured or semi-structured, and comes in different forms, such as images or texts. In order to process larger and larger amounts of data, researchers need to develop new techniques that can extract relevant information and infer some kind of structure from the available data. One of the principal domains that study such methods is machine learning, but there are many other related domains that aim to extract useful information from data. There are plenty of research studies~\cite{Hinton-NIPS-2012, Lazebnik-BBF-2006, Leung-2001, IIR-Manning-2008, Perronnin-ECCV-2010} in this direction. One of the main scopes of machine learning is to define a good representation of the data in order to build accurate classifiers. In text processing, implementing a simple \emph{bag of words} (BOW) model to represent a collection of documents can prove to be useful in tasks such as sentiment analysis~\cite{Pang-EMNLP-2002}, text categorization~\cite{Joachims-ECML-1998} or information retrieval~\cite{IIR-Manning-2008}. On the other hand, in order to process images, one should first find salient features before extracting them. The features can either be determined by experts in the specific domain of the application, or by a technique termed \emph{representation learning}, where the features are discovered automatically~\cite{Hinton-NIPS-2012,Bengio-2009, nn-tricks-2012}. As text documents, images can be represented using the bag of words model, but a \emph{word} has a completely different meaning and interpretation than in string and text processing. In fact, in computer vision, this model is known as the \emph{bag of visual words} (BOVW)~\cite{Csurka-2004, Sivic-VW-ICCV-2005, Zhang-LFK-2007}, and a \emph{visual word} is usually defined as a cluster of similar image descriptors~\cite{Dalal-HOG-2005, Lowe-SIFT-1999, Lowe-SIFT-2004, Bay-SURF-2008} extracted from the images.

In recent years, researchers have developed better ways~\cite{Mikolov-NIPS-2013} for representing words as vectors. Word embeddings~\cite{Mikolov-NIPS-2013,Bengio-JMLR-2003,Collobert-ICML-2008} have had a huge impact in natural language processing (NLP) and related fields, being used in many tasks including information retrieval~\cite{Perronnin-ACL-2013,Ye-ACM-2016}, sentiment analysis~\cite{Dos-COLING-2014} and word sense disambiguation~\cite{Chen-EMNLP-2014,Bhingardive-NAACL-2015,Navigli-ACL-2016,ShotgunWSD-EACL-2017}, among many others. In this paper, we look at word embeddings from a different perspective by drawing our inspiration from computer vision. Our aim is to redesign an efficient computer vision technique and use it for natural language processing tasks by leveraging the use of word embeddings. More specifically, we interpret word embeddings as \emph{text descriptors}, which allows us to adapt computer vision techniques based on local \emph{image descriptors} such as SIFT~\cite{Lowe-SIFT-1999, Lowe-SIFT-2004} or SURF~\cite{Bay-SURF-2008}. In computer vision, a \emph{local image descriptor} is a visual unit that represents a small image region by its elementary characteristics such as shape, color or texture. In natural language processing, word embeddings capture the semantic similarities between linguistic items and a \emph{text descriptor} (word vector) is a textual unit that represents a word by its semantic characteristics. Based on this analogy, we propose a novel approach for text classification inspired by the bag of visual words model. Our approach is different from the standard bag of words model used in natural language processing. Instead of using a vocabulary of words, we build a vocabulary of \emph{super word vectors} by clustering word vectors with k-means. Hence, a document will be represented as a \emph{bag of super word embeddings} (BOSWE). We also diverge from the idea of building a single vocabulary for the entire collection of documents, and propose to build a set of class-specific vocabularies of \emph{super word vectors}, by using the class labels earlier in the training process, in order to separate the samples before applying the k-means clustering. This latter approach seems to give better results in practice. In the learning stage, we employ kernel methods. We try out several kernels, such as the linear kernel, the intersection kernel, the Hellinger's kernel, the Jensen-Shannon kernel, and the relatively new PQ kernel~\cite{radu-PQK-ICIAP-2013, Ionescu_Popescu_PRL_2014}.
 
We underline that our contributions presented in this work are:
\begin{itemize}
\item an approach for adapting the bag of visual words~\cite{Csurka-2004,Sivic-VW-ICCV-2005,Zhang-LFK-2007} model from computer vision to natural language processing by leveraging the use of word embeddings;
\item an alternative approach for building a better representation based on class-specific vocabularies of super word embeddings;
\item an empirical evaluation demonstrating that the proposed model can obtain better results than a standard bag of words.
\end{itemize}

This paper is organized as follows. Section~\ref{sec_related_work} presents related work from computer vision and natural language processing. The bag of super word embeddings is presented in Section~\ref{sec_bowe}. We present experiments on polarity classification in Section~\ref{sec_polarity_classification} and on text categorization by topic in Section~\ref{sec_text_categorization}. Finally, we draw our conclusion in Section~\ref{sec_conclusion}.

\section{Related Work}
\label{sec_related_work}

\subsection{Bag of Visual Words}

Despite of the traditional view that computer vision and text processing are separate and unrelated fields of study, there are many cases in which text and images can be treated in a similar manner~\cite{radu-marius-book-2016}. One such example is the \emph{bag of words} representation. The bag of words model represents a text as an unordered collection of words, completely disregarding grammar, word order, and syntactic groups. It has many applications from information retrieval~\cite{IIR-Manning-2008} to natural language processing~\cite{FSN-Manning-1999} and word sense disambiguation~\cite{agirre-2006}. In the context of image analysis, the concept of \emph{word} had to be somehow defined. Certainly, computer vision researchers have introduced the concept of \emph{visual word} as described next. Local image descriptors, such as SIFT~\cite{Lowe-SIFT-1999, Lowe-SIFT-2004} or SURF~\cite{Bay-SURF-2008}, are vector quantized to obtain a vocabulary of visual words. The vector quantization process can be done, for example, by k-means clustering~\cite{Leung-2001} or by probabilistic Latent Semantic Analysis~\cite{Sivic-VW-ICCV-2005}. The frequency of each visual word is then recorded in a histogram which represents the final feature vector for the image. This histogram is the equivalent of the bag of words representation for text. The idea of representing images as \emph{bag of visual words} has demonstrated impressive levels of performance for image categorization~\cite{Zhang-LFK-2007}, image retrieval~\cite{Philbin-2007}, facial expression recognition~\cite{radu-WREPL-2013} and related tasks.

\subsection{Word Embeddings}

Because of the success of the bag of visual words model in image classification, we propose a similar approach on text, by replacing the local image descriptors with word embeddings. Word embeddings are well known in the NLP community~\cite{Bengio-JMLR-2003,Collobert-ICML-2008}, but they have recently become more popular due to the \emph{word2vec}~\cite{Mikolov-NIPS-2013} framework that allows to efficiently build vector representations from words. Word embeddings represent each word as a low-dimensional real valued vector, such that semantically related words reside in close vicinity in the generated space. Word embeddings are in fact a learned distributed representation of words where each dimension represents a latent feature of the word~\cite{Turian-ACL-2010}. Using the word representation induced by the embedding space, documents can be represented as a set of word vectors, where the size of this set is given by the number of words in the document. Given the fact that two documents are likely to be represented by sets of different sizes, the comparison between the respective documents cannot be done directly. To overcome this issue, Let et al.~\cite{Le-ICML-2014} propose the \emph{Paragraph Vector}, an unsupervised algorithm that learns fixed-length feature representations from variable-length pieces of texts, such as sentences, paragraphs, and documents. Their algorithm represents each document by a dense vector which is trained to predict words in the document. With some inspiration from computer vision, an alternative approach to solve the issue of variable-length representations is proposed by Clinchant et al.~\cite{Perronnin-ACL-2013}. Following the success of Fisher vectors in computer vision~\cite{Perronnin-ECCV-2010}, Clinchant et al.~\cite{Perronnin-ACL-2013} apply the Fisher kernel framework~\cite{Jaakkola-Haussler-1999} to aggregate the word embeddings of a document in order to obtain a fixed-length vector representation for the respective document. We propose a different approach that also draws its roots in computer vision research~\cite{Leung-2001,Csurka-2004,Philbin-2007}. Our approach employs the k-means clustering algorithm in order to group the word embeddings into a fixed number of clusters according their semantic relatedness. We regard the resulted cluster centroids as \emph{visual words} and process them accordingly, in order to obtain a histogram representation for each document.

Word embeddings have also been used in information retrieval~\cite{Perronnin-ACL-2013,Ye-ACM-2016} and in word sense disambiguation~\cite{Chen-EMNLP-2014,Bhingardive-NAACL-2015,Navigli-ACL-2016,ShotgunWSD-EACL-2017} due to their ability of modeling syntactic and semantic information. Another useful characteristic of word embeddings is that one can specifically train them to capture sentiment information in order to detect the polarity of documents~\cite{Dos-COLING-2014,Le-ICML-2014}.

\section{Bag of Super Word Embeddings}
\label{sec_bowe}

In computer vision, the BOVW model can be applied to image classification and related tasks, by treating image descriptors as words. A bag of visual words is a vector of occurrence counts of a vocabulary of local image features. This representation can also be described as a histogram of visual words. The vocabulary is usually obtained by vector quantizing image features into visual words.

Inspired by the BOVW model, we propose a similar way to process text documents by leveraging the use of word embeddings. In our approach designed for text, the image descriptors are replaced by word embeddings. Knowing the fact that word embeddings carry semantic information by projecting semantically related words in the same region of the embedding space, we propose to cluster word vectors in order to obtain relevant semantic clusters of words. Each centroid of the newly formed clusters can be regarded as a \emph{super word vector} that represents all the word vectors in a small region of the embedding space. By putting the super word vectors together, we obtain a vocabulary that we subsequently use to describe each document as a histogram of super word embeddings. We term this model \emph{bag of super word embeddings} (BOSWE).

\begin{figure*}[!tpb]
\begin{center}
\includegraphics[width=0.56\linewidth]{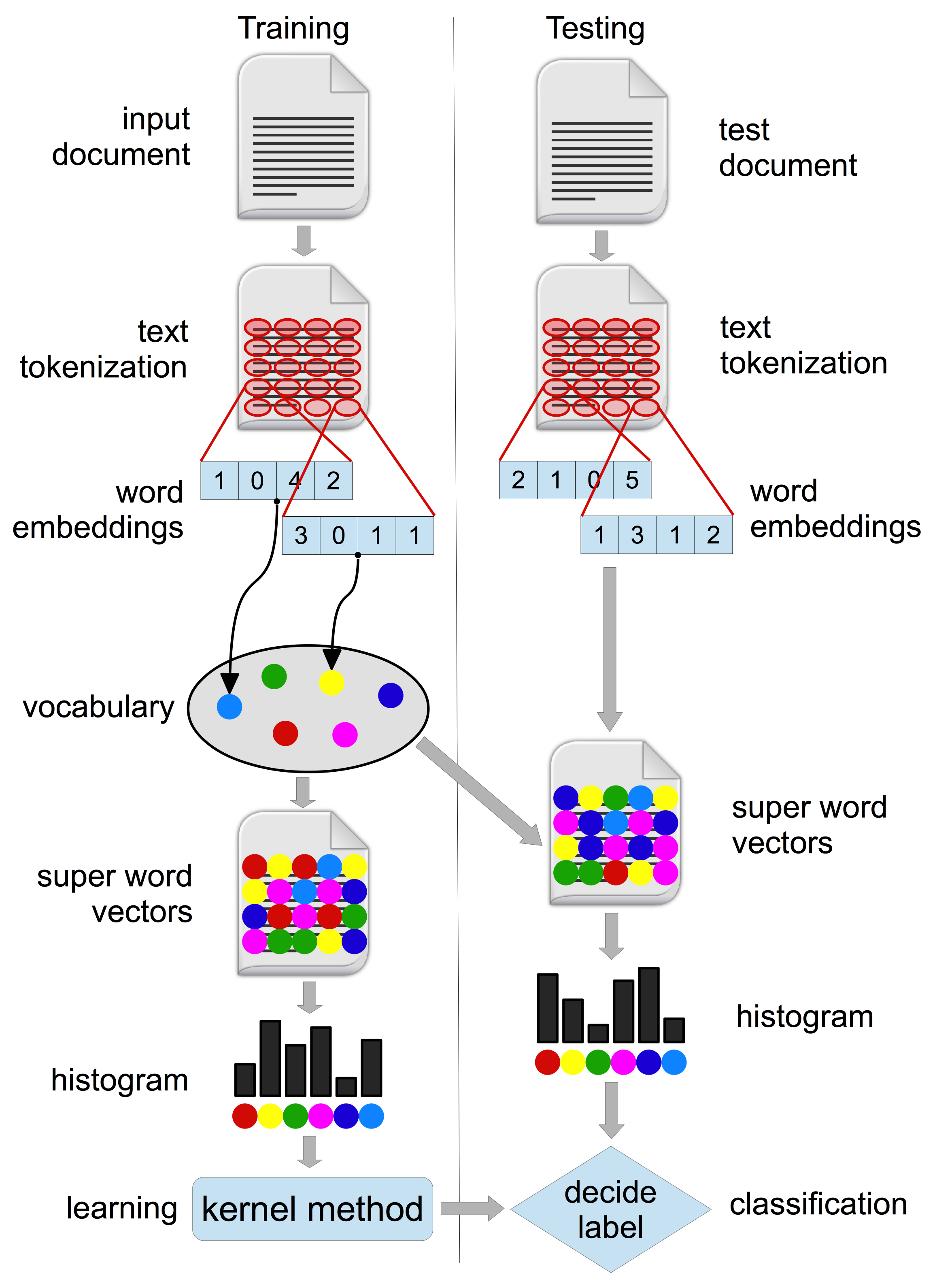}
\end{center}
\caption{The BOSWE model for text classification. Words are embedded into a vector space and quantized into super word vectors. The frequency of each super word vector is then recorded in a histogram. The histograms enter the training stage. Learning is done by a kernel method.}
\label{Fig_BOWE_Model}
\end{figure*}

The BOSWE model can be divided in two major steps. The first step is to build a feature representation. The second step is to train a kernel method in order to predict the class label of a new document. Each of these two steps are independently carried out in two stages, one for training (usually done offline) and one for testing (usually executed online). The entire process, that involves both training and testing stages, is illustrated in Figure~\ref{Fig_BOWE_Model}.

The feature representation step works as described next. Features are represented by the word vectors obtained by embedding all the words in the text. Next, the word embeddings are vector quantized and a vocabulary of super word embeddings is obtained. The vector quantization process is done by k-means clustering~\cite{Leung-2001}, and the formed centroids are stored in a randomized forest of k-d trees~\cite{Philbin-2007} to reduce search cost. Although alternative clustering approaches have been proposed in the computer vision literature~\cite{Sivic-VW-ICCV-2005,Martinet-VISAPP-2014}, k-means remains the most popular choice for the vector quantization step. This is the main reason for using k-means in our framework. The resulted centroids can hold some high-level abstract definition of a concept. Every word in the text is assigned to the closest centroid based on the Euclidean distance measure. The frequency of each super word embedding is then computed and recorded in a histogram. We propose two alternative pipelines for building the feature representation. In the first pipeline, we process the entire collection of documents all at once in order to build a single vocabulary of super word vectors. However, in this approach, words representing different classes can be clustered together due their semantic relatedness. The second pipeline aims to overcome this issue by grouping the training text documents into classes, and by processing each group of documents separately. This leads to a set of class-specific vocabularies of super word embeddings. Even if we know that a document belongs to a certain class at training time, we cannot rely on this assumption at test time. Therefore, we have to build the feature representation of each document by concatenating all the histograms corresponding to the class-specific vocabularies. For both pipelines, we then consider the feature vectors corresponding to the entire set of documents for the training step. Typically, a kernel method is employed for training the model. In computer vision, several kernels have be used at this stage. The linear kernel, the intersection kernel, the Hellinger's kernel or the Jensen-Shannon (JS) kernel are typical choices from the literature~\cite{Vedaldi-add-ker-2010}. Another option is the recently developed PQ kernel~\cite{radu-PQK-ICIAP-2013,Ionescu_Popescu_PRL_2014}. The underlying idea of the PQ kernel is to treat the super word vector histograms as ordinal data, in which data is ordered but cannot be assumed to have equal distance between values. In this case, a histogram will be regarded as a ranking of super word vectors according to their frequencies in that histogram. Usage of the ranking of super word vectors instead of the actual values of the frequencies may seem as a loss of information, but the process of ranking can actually make the PQ kernel more robust, acting as a filter and eliminating the noise contained in the values of the frequencies. As for object recognition~\cite{radu-PQK-ICIAP-2013,Ionescu_Popescu_PRL_2014} or texture classification in images~\cite{radu-andreea-marius-WSCG-2014}, we show that this kernel can yield better results than the other kernels. 

We try out the above mentioned kernel functions in combination with the Support Vector Machines (SVM) classifier~\cite{cortes-vapnik-ml-1995,taylor-Cristianini-cup-2004}. After our model is trained, it can be used to classify new documents. Given a test document, features are extracted and quantized into centroids from the vocabulary (or the multiple class-specific vocabularies) that was (were) already obtained in the training stage. The histogram of super word embeddings that represents the test document can be compared with the histograms learned in the training stage. The system can return either a label (or a score) for the test document or a ranked list of documents similar to the test document, depending on the application. For text classification a label (or a score) is enough, while for information retrieval a ranked list of documents is more appropriate. No matter the application, the training stage of the BOSWE model can be done offline. For this reason, the time that is necessary for vector quantization and learning is not of great importance. What matters most in the context of text classification is to return the result for a new (test) document as quickly as possible.

The performance level of the described model depends on the number of training documents, but also on the number of clusters. The number of clusters $k$ is a parameter of the model that must be set a priori. In computer vision, there is a common practice to use larger vocabularies from improved performance~\cite{Ionescu_Popescu_PRL_2014, radu-WREPL-2013}, however, there is a point where the accuracy saturates and the only effect of further increasing $k$ is to unnecessarily slow down the computation. 

\subsection{Implementation Details}

We next provide some implementation details for our BOSWE model used throughout the experiments. In the feature representation step, we have used the pre-trained word embeddings computed by the \emph{word2vec} toolkit~\cite{Mikolov-NIPS-2013} on the Google News data set using the Skip-gram model. The pre-trained model contains $300$-dimensional vectors for $3$ million words and phrases. Most of the steps involved in the BOSWE model, such as the k-means clustering and the randomized forest of k-d trees, are implemented using the VLFeat library~\cite{vedaldi-vlfeat-2008}. After computing the histograms, we apply one of the following kernels: the $L_2$-normalized linear kernel, the $L_1$-normalized Hellinger's kernel, the $L_1$-normalized intersection kernel, the $L_1$-normalized Jensen-Shannon (JS) kernel, and the $L_2$-normalized PQ kernel. The norms are chosen according to Vedaldi et al.~\cite{Vedaldi-add-ker-2010}, who state that $\gamma$-homogeneous kernels should be $L_{\gamma}$-normalized. We use the software provided by Ionescu et al.~\cite{Ionescu_Popescu_PRL_2014} to compute the PQ kernel. It is important to mention that all these kernels are used in the dual form, that implies using the \emph{kernel trick}~\cite{taylor-Cristianini-cup-2004} to directly build kernel matrices of pairwise similarities between samples. In the learning stage, we use the dual implementation of the Support Vector Machines classifier provided in LibSVM~\cite{LibSVM-2011}.

\section{Polarity Classification Experiments}
\label{sec_polarity_classification}

\subsection{Data Set}

The first corpus used to evaluate the proposed model is the Movie Review\footnote{{http://www.cs.cornell.edu/people/pabo/movie-review-data/}} data set~\cite{Pang-EMNLP-2002}. This is probably the most popular corpus used for sentiment analysis. The Movie Review data set consists of $2000$ movie reviews taken from the IMDB movie review archives. There are $1000$ positive reviews consisting of four and five star reviews, and $1000$ negative ones consisting of one and two star reviews. We use a $10$-fold cross-validation procedure in the evaluation. 

\begin{table*}[!t]
\caption{Accuracy rates using $10$-fold cross-validation on the Movie Review data set with different kernels and vocabulary dimensions. The best accuracy rate for each vocabulary dimension is highlighted in bold.}
\label{BOSWE_MR_ResultsTable}
\begin{center}
\begin{tabular*}{\hsize}{@{\extracolsep{\fill}}lccccc@{}}
\hline
Vocabulary 	& Linear ($L_2$-norm)	& Hellinger's ($L_1$-norm)	& Intersection ($L_1$-norm) & JS ($L_1$-norm) & PQ ($L_2$-norm) \\
\hline
$1 \times 5000$ words 		&$84.80\%$		&$86.15\%$		&$85.40\%$ 		&$85.80\%$ 		&$\mathbf{86.55\%}$\\
$1 \times 10000$ words 	&$85.05\%$		&$86.45\%$		&$85.75\%$		&$86.10\%$		&$\mathbf{87.15\%}$\\
$2 \times 5000$ words 		&$85.75\%$		&$87.60\%$		&$86.95\%$		&$87.35\%$ 		&$\mathbf{88.25\%}$\\	
$2 \times 7500$ words		&$87.15\%$ 		&$88.60\%$		&$88.15\%$		&$87.80\%$ 		&$\mathbf{88.95\%}$ \\
\hline
\end{tabular*}
\end{center}
\end{table*}

\begin{table*}[!t]
\caption{Accuracy rates using $10$-fold cross-validation on the Movie Review data set with various BOSWE configurations versus two baseline approaches. The best accuracy rate is highlighted in bold.}
\label{MR_ResultsTable}
\begin{center}
\begin{tabular*}{\hsize}{@{\extracolsep{\fill}}lc@{}}
\hline
Method 													& Accuracy \\
\hline
Baseline BOW																					& $84.10\%$ \\
Pang et al.~\cite{Pang-EMNLP-2002}												& $82.90\%$ \\
BOSWE ($2 \times 7500$ words and Hellinger's kernel) 				& $88.60\%$ \\
BOSWE ($2 \times 7500$ words and PQ kernel) 							& $88.95\%$ \\
BOSWE ($2 \times 7500$ words and Hellinger's kernel $+$ PQ kernel) 	& $\mathbf{89.65\%}$ \\
\hline
\end{tabular*}
\end{center}
\end{table*}

\subsection{Baselines}

We compare our model against a baseline bag of words. We considered the following steps to obtain a bag of words representation suited for the polarity categorization task. First of all, the text is broken down into tokens. After applying the tokenization process, the next step is to eliminate the stop words\textsuperscript{(}\footnote{Stop words are the most common words in a language, usually function words, such as \emph{this}, \emph{is}, \emph{it}.}\textsuperscript{)}. The remaining terms from the entire collection of documents are gathered into a vocabulary. The frequency of each term is then computed on a per document basis. The frequency histograms are normalized using the $L_2$-norm. As in our own approach, we use SVM for training. We also consider the approach of Pang et al.~\cite{Pang-EMNLP-2002}, an alternative implementation of the bag of words model, as baseline.

\subsection{Results}

Table~\ref{BOSWE_MR_ResultsTable} presents the accuracy rates of various BOSWE models obtained in a $10$-fold cross-validation procedure carried out on the Movie Review data set, by combining different vocabulary dimensions and kernels. The results presented in Table~\ref{BOSWE_MR_ResultsTable} indicate that building a vocabulary for each polarity class (positive and negative) is a better approach than building a single vocabulary for the entire training set. This observation holds for every kernel considered in the evaluation. Interestingly, among the evaluated kernels, we obtain better performance with the Hellinger's and the PQ kernels. For every vocabulary dimension, PQ kernel always yields the best results. The best performance ($88.95\%$) is obtained when the BOSWE model relies on two vocabularies, each of $7500$ super word vectors, and on the PQ kernel. Remarkably, these results are somewhat consistent to the results reported by Ionescu et al.~\cite{radu-PQK-ICIAP-2013,Ionescu_Popescu_PRL_2014} in the context of object recognition from images. Indeed, Ionescu et al.~\cite{radu-PQK-ICIAP-2013,Ionescu_Popescu_PRL_2014} have also found that using more visual words and applying the PQ kernel leads to better performance.

We compare our best BOSWE configurations with two baseline approaches in Table~\ref{MR_ResultsTable}. We also try to combine the Hellinger's and the PQ kernels by summing them up, in order to improve the performance. Nevertheless, the results indicate that all our BOSWE configurations achieve better performance than the baseline approaches. The best BOSWE configuration yields an accuracy of $89.65\%$. Our best approach is more $5\%$ better than baseline BOW and more than $6\%$ better than the baseline approach of Pang et al.~\cite{Pang-EMNLP-2002}. We thus conclude that the BOSWE model is capable to improve the performance over a standard BOW model for the polarity classification task.

\section{Text Categorization Experiments}
\label{sec_text_categorization}

\subsection{Data Set}

The Reuters-21578\footnote{{http://www.daviddlewis.com/resources/testcollections/reuters21578/}} corpus~\cite{reuters-21578} is one of the most widely used test collections for text categorization research. It contains $21578$ articles collected from Reuters newswire. Following the procedure of Joachims et al.~\cite{Joachims-ECML-1998} and Yang et al.~\cite{Yang-SIGIR-1999}, the categories that have at least one document in the training set and one in the test set are selected. This leads to a total of $90$ categories. We use the ModeApte evaluation~\cite{Xue-TKDE-2009}, in which unlabeled documents are eliminated. After removing the unlabeled documents, there are $10787$ documents left that belong to $90$ categories. Each document belongs to one or more categories and the average number of categories per document is $1.2$. The collection is split into $7768$ documents in the training set and $3019$ documents in test set.

\subsection{Baseline}

We compare our BOSWE model with a bag of words baseline adapted specifically to text categorization by topic. The following steps are required to obtain a bag of words representation suited for the text categorization task. The text is first broken down into tokens. After tokenization, the following step is to eliminate the stop words, as they do not provide useful information in the context of text categorization by topic. The remaining words are stemmed using the Porter stemmer~\cite{Porter-1980} algorithm\textsuperscript{(}\footnote{Stemming is the process that reduces a word to its root form.}\textsuperscript{)}. This algorithm removes the commoner morphological and inflexional endings from words in English. The resulted terms from the entire collection of documents are collected into a vocabulary. The frequency of each term is then computed on a per document basis. Let $f_{t,d}$ denote the raw frequency of a term $t$ in a document $d$, namely the number of times $t$ occurs in $d$. The bag of words representation used as baseline in the following experiments is obtained by computing the log normalized \emph{term frequency} as follows:
\begin{equation}\label{eq_log_tf}
\begin{split}
tf(t,d) = \left\{\begin{array}{ll} 1 + log f_{t,d}, \; \mbox{if} \; f_{t,d} > 0 \\ 0, \; \mbox{if} \; f_{t,d} = 0\end{array}\right. .
\end{split}
\end{equation}



\subsection{Evaluation Procedure}

To evaluate and compare the text categorization approaches, the precision and the recall are first computed based on the confusion matrix presented in Table~\ref{Tab_confusion_pr_rec}. The \emph{precision} is given by the number of true positive documents ($TP$) divided by the number of documents predicted as positive by the classifier ($TP+FP$), while the \emph{recall} is given by the number of true positive documents ($TP$) divided by the total number of documents marked as positive by a trusted expert judge ($TP+FN$). To capture the precision and recall into a single representative number, the $F_1$ measure can be employed. The $F_1$ measure can be interpreted as a weighted average of the precision and recall given by:
\begin{equation*}
F_{1} = 2 \cdot \frac{precision \cdot recall}{precision + recall}.
\end{equation*}

\begin{table*}[!t]
\small{
\begin{center}
\begin{tabular}{|l|c|c|c|}
\hline
\multicolumn{2}{|c|}{}	& \multicolumn{2}{c|}{Expert judgments} \\
\cline{2-4}
							&	Labels	& $+1$			& $-1$\\
\hline
Classifier				& $+1$	& $TP$				& $FP$ \\
\cline{2-4}
predictions			& $-1$		& $FN$				& $TN$ \\
\hline
\end{tabular}
\end{center}
\caption{Confusion matrix of a binary classifier with labels $+1$ or $-1$. There are four distinct groups of samples illustrated here: true positive ($TP$), false positive ($FP$), false negative ($FN$), and true negative ($TN$).}
\label{Tab_confusion_pr_rec}
}
\end{table*}

For each category, a binary classifier is trained to predict the positive and negative labels for the test documents. However, the performance of the classifier needs to be evaluated at the global level (over all categories). Two approaches are used in literature to aggregate the $F_1$ measures over multiple categories. One is based on computing a confusion matrix for each category, which can be used to subsequently calculate the $F_1$ measure for each category. Finally, the global $F_1$ measure is obtained by averaging all the $F_1$ measures. This first measure is known as macro-averaged $F_1$ ($macroF_1$). The other approach is based on computing a global confusion matrix for all the categories by summing the documents that fall in each of the four conditioned sets, namely true positives, true negatives, false positives, and false negatives. The global $F_1$ measure is immediately computed with the values provided by the global confusion matrix. This second measure is known as micro-averaged $F_1$ ($microF_1$). As noted by Xue et al.~\cite{Xue-TKDE-2009}, the classifier's performance on rare categories has more impact on the macro-averaged $F_1$ measure, while the performance on common categories has more impact on the micro-averaged $F_1$ measure. Thus, it makes sense to report both these measures in the following experiments.

\subsection{Results}

\begin{table*}[!t]
\scriptsize{
\caption{Accuracy rates on the Reuters-21578 test set with different kernels and vocabulary dimensions. The best $microF_1$ and $macroF_1$ scores for each vocabulary dimension are highlighted in bold.}
\label{BOSWE_Reu_ResultsTable}
\begin{center}
\begin{tabular*}{\hsize}{@{\extracolsep{\fill}}lccccclccccc@{}}
\hline
Vocabulary 	& \multicolumn{2}{c}{Linear ($L_2$-norm)}	& \multicolumn{2}{c}{Hellinger's ($L_1$-norm)}	& \multicolumn{2}{c}{Intersection ($L_1$-norm)} & \multicolumn{2}{c}{JS ($L_1$-norm)} & \multicolumn{2}{c}{PQ ($L_2$-norm)} \\
 & $microF_1$ & $macroF_1$ & $microF_1$ & $macroF_1$ & $microF_1$ & $macroF_1$ & $microF_1$ & $macroF_1$ & $microF_1$ & $macroF_1$ \\
\hline
$1 \times 10000$ words 	&$86.62\%$	&$\mathbf{49.42\%}$	&$86.56\%$ 	& $45.21\%$		&$85.28\%$ 	& $41.19\%$		&$86.30\%$ 	& $43.30\%$		&$\mathbf{86.74\%}$ & $49.31\%$\\

$1 \times 20000$ words 	&$86.72\%$	&$\mathbf{49.58\%}$	&$86.61\%$ 	&$45.39\%$		&$85.66\%$	&	$41.55\%$	&$86.35\%$	& $43.54\%$		&$\mathbf{86.80\%}$ &$49.36\%$\\

$90 \times 100$ words 		&$86.77\%$	&$\mathbf{49.63\%}$	&$86.91\%$	&$47.71\%$		&$86.25\%$	&$42.50\%$		&$86.59\%$ 	&$44.94\%$		&$\mathbf{86.84\%}$ & $49.49\%$\\	

$90 \times 200$ words		&$86.83\%$ 	&$\mathbf{49.68\%}$	&$87.04\%$	&$47.75\%$		&$86.33\%$	&$42.64\%$		&$86.74\%$ 	&$45.06\%$		&$\mathbf{87.07\%}$ & $49.51\%$ \\
\hline
\end{tabular*}
\end{center}
}
\end{table*}

Table~\ref{BOSWE_Reu_ResultsTable} presents the micro-averaged $F_1$ scores and macro-averaged $F_1$ scores of various BOSWE models obtained on the Reuters-21578 test set, by combining different vocabulary dimensions and kernels. The results presented in Table~\ref{BOSWE_Reu_ResultsTable} indicate that building a vocabulary for each topic gives slightly better results than building a single vocabulary for all the $90$ topics, even though the topic-specific vocabularies are significantly smaller in size, e.g. $200$ words versus $20000$ words. Among the evaluated kernels, we obtain better performance with the linear and the PQ kernels. While the PQ kernel yields a better $microF_1$ score, the linear kernel compensates with a better $macroF_1$ score. Nonetheless, the difference between the two kernels is not significant.

\begin{table*}[!t]
\caption{Accuracy rates on the Reuters-21578 test set with various BOSWE configurations versus a baseline bag of words model. The best $microF_1$ and $macroF_1$ scores are highlighted in bold.}
\label{Reu_ResultsTable}
\begin{center}
\begin{tabular*}{\hsize}{@{\extracolsep{\fill}}lcc@{}}
\hline
Method 																							& $microF_1$ 					& $macroF_1$ \\
\hline
Baseline BOW																					& $86.09\%$ 					& $49.45\%$\\
BOSWE ($90 \times 200$ words and linear kernel) 						& $86.83\%$ 					& $49.68\%$\\
BOSWE ($90 \times 200$ words and PQ kernel) 							& $87.07\%$ 					& $49.51\%$\\
BOSWE ($90 \times 200$ words and linear kernel $+$ PQ kernel) 			& $\mathbf{87.24\%}$ 	& $\mathbf{49.72\%}$\\
\hline
\end{tabular*}
\end{center}
\end{table*}

We compare our best BOSWE configurations with two baseline approaches in Table~\ref{Reu_ResultsTable}. We again try to combine best performing kernels by summing them up. Although the results indicate that our BOSWE configurations achieve better performance than the baseline bag of words, the differences are not as high as in the polarity classification experiments. Our best BOSWE configuration yields a $microF_1$ score of $87.24\%$ and  a $macroF_1$ score of $49.72\%$, which represents an improvement of $1.15\%$ in terms of $microF_1$ and $0.27\%$ in terms of $macroF_1$ over the baseline. Overall, it seems that the BOSWE model can surpass the performance of a standard bag of words representation for text categorization by topic.

\section{Conclusion}
\label{sec_conclusion}

In this paper, we have presented an approach for building an effective feature representation for various text classification tasks. The proposed approach is based on clustering word embeddings using k-means and on representing a text document as a \emph{bag of super word embeddings}, in a similar fashion to the \emph{bag of visual words} model, which is broadly used in computer vision for representing images. The empirical results on polarity classification and text categorization by topic demonstrate that our approach is able to surpass the classical \emph{bag of words} approach. 

Some researchers~\cite{Martinet-VISAPP-2014} have questioned the suitability of the k-means algorithm for the vector quantization of visual words, as the generated clusters (visual words) do not follow Zipf's law, although words in natural language do follow it. In future work, we aim to replace the k-means clustering approach with alternative approaches, such as density-based clustering or self-organizing maps. In case the words projected in the embedding space are not uniformly distributed, it would be more appropriate to employ a clustering algorithm that is able to capture the distribution of the embedded words into a vocabulary that follows Zipf's law. According to Martinet~\cite{Martinet-VISAPP-2014}, this can lead to more accurate results.

\section*{Acknowledgements}

The authors have equally contributed to this work. The authors thank the reviewers for their helpful comments.




{\small
\bibliographystyle{ieee}
\bibliography{references}
}



\end{document}